\documentclass{article}
\usepackage{amsmath,graphicx,amssymb,bm,amsthm,subcaption}
\usepackage{algorithmic,algorithm}
\usepackage{balance,cite,color}

\usepackage[preprint]{spconf}
\copyrightnotice{\begin{minipage}{\textwidth}\footnotesize\copyright \ 
	20XX IEEE. Personal use of this material is permitted. Permission from IEEE must be obtained for all other uses, in any current or future media, including reprinting/republishing this material for advertising or promotional purposes, creating new collective works, for resale or redistribution to servers or lists, or reuse of any copyrighted component of this work in other works. \end{minipage} }

\newcommand{\grad}[1]{\nabla_{#1^{\tran}} }
\newcommand{\w}{\bm{w}}

\newcommand{\we}{\widetilde{\w}}
\newcommand{\eqdef}{\:\overset{\Delta}{=}\:}
\DeclareMathOperator*{\argmin}{argmin}
\newcommand{\Li}{\mathcal{L}_i}
\newcommand{\B}[2]{\mathcal{B}_{#1,i,#2}}
\newcommand{\tran}{{\sf T}}

\newtheorem{theorem}{Theorem}
\newtheorem{assumption}{Assumption}

\title{Optimal Importance Sampling for Federated Learning}
\name{Elsa Rizk, Stefan Vlaski, Ali H. Sayed
\thanks{School of Engineering, École Polytechnique Fédérale de Lausanne, 1015 Lausanne, Switzerland (e-mail:\{elsa.rizk, stefan.vlaski, ali.sayed\}@epfl.ch).}}
\address{\vspace{-2cm}}
\begin{document}
\ninept
\maketitle
\begin{abstract}
Federated learning involves a mixture of centralized and decentralized processing tasks, where a server regularly selects a sample of the agents and these in turn sample their local data to compute stochastic gradients for their learning updates. This process runs continually. The sampling of both agents and data is generally uniform; however, in this work we consider non-uniform sampling. We derive optimal importance sampling strategies for both agent and data selection and show that non-uniform sampling without replacement improves the performance of the original FedAvg algorithm. We run experiments on a regression and classification problem to illustrate the theoretical results. 

\end{abstract}
\begin{keywords}
federated learning, importance sampling, asynchronous SGD, non-IID data, heterogeneous agents
\end{keywords}
\vspace{-0.1cm}
\section{Introduction and Related Material}
\label{sec:intro}
\vspace{-0.1cm}
Most current machine learning applications work on large datasets, and sometimes this data is distributed accross several locations, making learning algorithms computationaly and communicatively expensive. Therefore, designers of these learning algorithms move toward stochastic implementations where only a subset of the data is used, such as in stochastic gradient descent (SGD). The most prominent selection schemes use uniform sampling. While it has proven to be a good approximator of the original solution, a greater advantage exists in non-uniform sampling that prioritize samples and result in a faster learning process. For SGD, importance sampling attributes to each data sample a probability proportional to the gradient evaluated at that sample. Such probabilities result in faster convergence and reduced stochastic gradient variance \cite{impSampSayed,alain2015variance,Nati14, zhao2015stochastic, jaggi17}. 
	
A prominent machine learning architecture is federated learning \cite{mcmahan16} where non-IID data is distributed accross heterogeneous agents, and a server acts as a master node and controller of the learning process. The goal is to find a model that fits all the data across the agents. If we consider $K$ agents each with a non-IID dataset $\{x_{k,n}\}$ of size $N_k$, where subscript $k$ denotes the agent and subscript $n$ indicates the sample number, the global optimization problem becomes as expressed in \eqref{eq:globalProb}, where the cost function $P_k(\cdot)$ is defined as the sample average of some loss function $Q_k(\cdot;x_{k,n})$. 
\vspace{-0.2cm}
\begin{equation}\label{eq:globalProb}
	w^o \eqdef \argmin_{w}  \frac{1}{K}\sum_{k=1}^K P_k(w)\eqdef \frac{1}{N_k} \sum_{n=1}^{N_k}Q_k(w;x_{k,n})
\end{equation}
In standard federated learning, at each iteration $i$ the server selects $L$ agents, and we denote by $\Li$ the set of indices of the chosen agents. In turn, each participating agent runs $E_k$ epochs, during which it selects a mini-batch $\B{k}{e}$ of size $B_k$.  The sampling occurs uniformly with replacement. Thus, unlike standard learning algorithms, two-level sampling exists in federated learning: first at the agent level and second at the data level. 

In the literature on federated learning, researchers have studied the effects of non-uniform selection of agents, i.e., the selection of agents with unequal probabilities. The selection schemes can be split into those maximizing accuracy and those prioritizing fairness. Of the works maximizing accuracy, the authors in \cite{nishio2019client} introduce a new agent selection scheme called FedCS. The selection protocol consists of two steps. In the first step, the server asks the available agents for information on their resources. In the second step, based on the recieved information, the server selects the most number of agents that are capable of finishing an interation by a set deadline. Reference \cite{HybridFL} broadens the previous work to include non-IID data, and they allow the server to act as one of the agents; the server collects from the selected agents some of their data and runs SGD on them. In \cite{nguyen2020fast}, a non-uninform sampling scheme of agents is considered; the probability distribution of the sampling of agents is calculated by maximizing the inner product between the global and local gradients. They provide an approximate solution, since the actual calculations of the sampling probabilities is non-trivial. In \cite{mohri19a,Li2020Fair}, the authors goal is to ensure fairness. In \cite{mohri19a} the authors introduce agnostic federated learning where they model the data distribution as a non-uniform mixture of the local distributions. Thus, they solve a minimax problem to select the agents. Such a solution leads to always minimizing the worst loss function. Building on the previous work, q-FFL \cite{Li2020Fair} reweights the cost function by assigning higher weights to agents that have larger loss. 

None of these works appears to consider the problem of incorporating importance samples at two levels: one for selecting the subset of agents and another for the agents to sample their own local data.  In what follows, we implement non-uniform sampling of agents and data without replacement. To do so, we first introduce the notion of \textit{normalized inclusion probability}. We define $p_k$ as the normalized inclusion probability of agent $k$ and $p_n^{(k)}$ as the normalized inclusion probability of data sample $n$ of agent $k$ \cite{horvitz1952generalization}:
	\vspace{-0.1cm}
\begin{equation}
	p_k \eqdef \frac{\mathbb{P}(k \in \Li)}{L}, \qquad p_n^{(k)} \eqdef \frac{\mathbb{P}(n \in \B{k}{e})}{B_k}.
\end{equation}
We briefly mention the difference between the \textit{inclusion probabilities} and the \textit{sampling probabilities}, since it is important to understand the distinction for the remainder of this work. The former indicates the total probability of a sample being chosen after the sampling scheme is done, while the latter is used to choose the samples. The sampling probability is not affected by the sampling scheme, while the inclusion probability is. When sampling with replacement, the two are equal up to a multiplicative factor. When sampling without replacement, a non-trivial relationship exists between them. For instance, if we require to sample 2 out of 4 balls with sampling probabilitis denoted by $\pi_n$, the normalized inclusion probability of ball 1 is given by: 
\vspace{-0.2cm}
\begin{align}
	&p_1 =\frac{1}{2} \sum_{n=2}^4 \mathbb{P}\big(\text{1 chosen on the $1^{st}$ trial \& } n \text{ chosen on the $2^{nd}$ trial}\big) \notag \\ 
	&\qquad\quad\:\:\:+\mathbb{P}\big(\text{$n$ chosen on the $1^{st}$ trial \& } \text{1 chosen on the $2^{nd}$ trial}\big) \notag \\
	&= \frac{1}{2}\sum_{n=2}^4 \pi_1  \frac{\pi_n}{1-\pi_1}+ \pi_n \frac{\pi_1}{1-\pi_n}.
\end{align}
 In standard importance sampling, we optimize over the sampling probabilities, while in our implementation we optimize over the normalized inclusion probabilities, since we sample without replacement.
\vspace{-0.4cm}
\section{Algorithm Description}
We introduce the ISFedAvg algorithm, which is a variation of the original FedAvg \cite{mcmahan16}. The goal of the original algorithm is to approximate the true gradient by some stochastic gradient. During each iteration $i$ of FedAvg algorithm, agent $k$ runs $E_k$ steps of SGD:
\vspace{-0.1cm}
\begin{align}\label{eq:localUp}
	\w_{k,e} = \w_{k,e-1} - \mu \frac{1}{B_k}\sum_{b \in \B{k}{e}} \grad{w} Q_k(\w_{k,e-1};\bm{x}_{k,b}),
\end{align}
where $\w_{k,0} = \w_{i-1}$. After each agent finishes its epochs, the server combines the new parameter vector as follows:
\vspace{-0.1cm}
\begin{align}\label{eq:modelAggregation}
	\w_{i} = \frac{1}{L}\sum_{k \in \Li} w_{k,E_k}.
\end{align}
The solution provided by FedAvg algorithm leads to the following estimation of the gradient at any given iteration $i$:
\begin{equation}
	 \frac{1}{L}\sum_{k \in \Li} \frac{1}{B_k}\sum_{e=1}^{E_k} \sum_{b \in \B{k}{e}} \grad{w} Q_k(\w_{k,e-1}; \bm{x}_{k,b}).
\end{equation}
However, as argued in \cite{DFL}, we modify the estimator by scaling it by the epoch size, so that the resulting estimator is unbiased. Furthermore, under more general sampling schemes, the unbiased estimator is in fact \cite{impSampSayed}:
\begin{equation}\label{eq:gradEst}
	\frac{1}{L}\sum_{k \in \Li}\frac{1}{Kp_k} \frac{1}{E_kB_k}\sum_{e=1}^{E_k} \sum_{b \in \B{k}{e}} \frac{1}{N_kp_b^{(k)}}\grad{w} Q_k(\w_{k,e-1}; \bm{x}_{k,b}).
\end{equation} 
Thus, the new algorithm operates as follows: At each iteration $i$, the server selects $L$ participating agents such that each agent's probability of inclusion is $\mathbb{P}(k \in \Li) = Lp_k$. Then, each agent $k$ initializes its model with the previous model sent by the server $\w_{i-1}$ and performs $E_k$ SGD steps. During each step, agent $k$ chooses its mini-batch such that a data point's probability of inclusion is $\mathbb{P}(b \in \B{k}{e}) = B_k p_b^{(k)}$. It then uses the following stochastic gradient to update its local model: 
\begin{equation}
	 \frac{1}{Kp_kE_kB_k} \sum_{b \in \B{k}{e}} \frac{1}{N_k p_b^{(k)}} \grad{w}Q_k(w; \bm{x}_{k,b}).
\end{equation}
After all the agents are done and send back their new model, the server aggregates them according to \eqref{eq:modelAggregation}. We call this new algorithm Importance Sampling FedAvg (ISFedAvg). 

\begin{algorithm}
	\begin{algorithmic}
		\caption{(Importance Sampling Federated Averaging)}\label{alg:AFL}
		\STATE{
			\textbf{initialize} $w_{0}$\;}
		\FOR{each iteration $i=1,2,\cdots$}\STATE{
			Select the set of participating agents $\Li$ by sampling \( L \) times from \( \{ 1, \ldots, K \} \) without replacement according to the probabilities $p_k$.\\
			\FOR{each agent $k \in \mathcal{L}_i$} \STATE {
				\textbf{initialize} $\w_{k,0} = \w_{i-1}$ \\
				\FOR{each epoch $e=1,2,\cdots E_{k}$}\STATE{
					Find indices of the mini-batch sample \( \B{k}{e} \) by sampling \( B_{k} \) times from \( \{ 1, \ldots, N_{k} \} \) without replacement according to the probabilities $p_n^{(k)}$.\\
					$\bm{g}  =\dfrac{1}{K p_{k}E_{k}B_k} \sum\limits_{b\in \B{k}{ e}} \dfrac{1}{N_{k}p_b^{(k)}}\grad{w}Q_{k}(\w_{k,e-1};\bm{x}_{k,b})$ \\
					$\w_{{k},e} = \w_{{k},e-1} - \mu\bm{g}$ \\
				}\ENDFOR
			} \ENDFOR
			\\ $\w_i = \dfrac{1}{L}\sum\limits_{k \in \Li} \w_{k,E_{k}}$
		}\ENDFOR
		
	\end{algorithmic}
\end{algorithm}

\section{Algorithm Convergence}
\vspace{-0.1cm}
We now examine the convergence behavior of the proposed ISFedAvg implementation. The analysis will suggest ways to optimize the selection of the sampling probabilities which we discuss in the next section.

\subsection{Modeling Conditions}
Certain standard assumptions on the nature of the cost functions and the local optimizers must be made to ensure a tractable convergence analysis. Thus, in this work we assume convexity of the cost functions and smoothness of their gradients. 
\begin{assumption}\label{assum:conLip}
	The functions $P_k(\cdot)$ are $\nu-$strongly convex, and $Q_k(\cdot;x_{k,n})$ are convex, namely:
	\begin{align}
		&P_k(w_2) \geq P_k(w_1) + \grad{w}P_k(w_1)(w_2-w_1) + \frac{\nu}{2}\Vert w_2-w_1\Vert^2, \\
		&Q_k(w_2;x_{k,n}) \geq Q_k(w_1;x_{k,n}) + \grad{w}Q_k(w_1;x_{k,n})(w_2-w_1).
	\end{align}
	Also, the functions $Q_k(\cdot; x_{k,n})$ have $\delta-$Lipschitz gradients: 
	\begin{align}
		\Vert \grad{w}Q_k(w_2;x_{k,n})-\grad{w}Q_k(w_1;x_{k,n})\Vert &\leq \delta \Vert w_2-w_1\Vert.
	\end{align}
\qed
\end{assumption}
\noindent
We further require the local optimizers $w_k^o = \argmin_w P_k(w)$ not to drift too far from the global optimizer. Such an assumption is necessary for good performance since without it, agents would have very different local models. The resulting average model would not perform well locally. 
\begin{assumption}\label{assum:bdLocalMin}
	The distance of each local model $w^o_k$ to the global model $w^o$ is bounded unifromly: 
	\begin{equation}
		\Vert w^o_k - w^o \Vert \leq \xi.
	\end{equation}
\qed
\end{assumption}
\vspace{-0.7cm}
\subsection{Main Convergence Result}
Instead of writing the recursion as in \eqref{eq:localUp} and \eqref{eq:modelAggregation} and using the more general estimator \eqref{eq:gradEst}, we can combine the multiple epochs and the aggregation into one SGD step:
\vspace{-0.2cm}
\begin{align}
	\w_i = \w_{i-1}  
	- \mu \frac{1}{L} &\sum_{k \in \Li} \frac{1}{Kp_kE_kB_k}\sum_{e=1}^{E_k}\sum_{b \in \B{k}{e}} \frac{1}{N_kp_b^{(k)}} \notag \\
	&\times \grad{w}Q_k(\w_{k,e-1};\bm{x}_{k,b}).
\end{align}
For simplicity of notation, we introduce the stochastic gradient:
\vspace{-0.2cm}
\begin{equation}
	\widehat{\grad{w}P_k}(w) \eqdef \frac{1}{E_kB_k}\sum_{e=1}^{E_k} \sum_{b \in \B{k}{e}} \frac{1}{N_k p_b^{(k)}} \grad{w}Q_k(w; \bm{x}_{x,b}).
\end{equation}
Then, by introducing $\we_i = w^o - \w_i$ and defining two error terms: 
\begin{align}
	\bm{s}_i \eqdef & \frac{1}{L} \sum_{k \in \Li}\frac{1}{Kp_{\ell}} \widehat{\grad{w}P_k}(\w_{i-1}) - \frac{1}{K} \sum_{k=1}^K \grad{w}P_k(\w_{i-1}), \label{eq:gradNoise} \\
	\bm{q}_i \eqdef & \frac{1}{L} \sum_{\ell \in \Li} \frac{1}{K p_{\ell}E_{\ell}B_{\ell}} \sum_{e=1}^{E_{\ell}}\sum_{b\in\B{\ell}{e}} \frac{1}{N_{\ell} p_{b}^{(\ell)}}	\notag \\
	& \times \big( \grad{w}Q_k(\w_{\ell,e-1};\bm{x}_{\ell,b})  - \grad{w}Q_k(\w_{i-1};\bm{x}_{\ell,b})\big) \label{eq:incNoise},
\end{align}
we can write the following error recursion:
\begin{align}\label{eq:errRec}
	\we_i = \we_{i-1} +\mu \bm{s}_i + \mu \bm{q}_i + \mu \frac{1}{K} \sum_{k=1}^K \grad{w}P_k(\w_{i-1}).
\end{align}
We call the first error term \textit{gradient error} \eqref{eq:gradNoise}, which quantifies the error due to the approximation of the gradient by using subsets of agents and data; it is the error that results when central SGD is ran at the server. We call the second error term \textit{incremental error} \eqref{eq:incNoise}, since it captures the error induced by running multiple epochs locally. We show that both errors have bounded second order moments; furthermore, we show that the gradient noise is zero-mean, which implies that the gradient estimate is unbiased. We summarize the results in the below two theorems. 

\begin{theorem}[\textbf{Estimation of first and second order moments of the gradient noise}]\label{thrm:gradNoise}
	The gradient noise $\bm{s}_i$ defined in \eqref{eq:gradNoise} is zero-mean	
	with bounded variance:
	\begin{align}\label{eq:thrm-bd-var-wo}
	\mathbb{E}\Vert \bm{s}_i\Vert^2\leq &  \beta_s^2 \mathbb{E}\Vert \widetilde{\w}_{i-1}\Vert ^2  + \sigma_s^2,
\end{align}
	where:
	\begin{align}\label{eq:thrm-cst-wo}
		\beta_s^2 \eqdef & 3\delta^2 + \frac{1}{K^2}\sum_{k=1}^K \frac{1}{p_k} \left( \beta_{s,k}^2 + 3\delta^2 \right) ,\\
		\sigma_s^2 \eqdef & \frac{1}{K^2}\sum_{k=1}^K \frac{1}{p_k}\left\{ \sigma_{s,k}^2 + \left( 3 + \frac{6}{E_kB_k}\right) \Vert \grad{w}P_k(w^o)\Vert^2 \right\}  , \\
		\beta_{s,k}^2 & \eqdef  \frac{3\delta^2}{E_kB_k} \left( 1 + \frac{1}{N_k^2}\sum_{n=1}^{N_k} \frac{1}{p_n^{(k)}} \right), \label{eq:locCstBeta} \\
		\sigma_{s,k}^2 & \eqdef \frac{6}{E_kB_kN_k^2}\sum_{n=1}^{N_k}\frac{1}{p_n^{(k)}} \Vert \grad{w}Q_k(w^o;x_{k,n}) \Vert^2 \label{eq:locCstSig}.
	\end{align}
\end{theorem}
\begin{proof}
	Proof omitted due to space limitations. 
\end{proof}
\noindent We observe that the gradient noise is controlled by the normalized inclusion probabilities. The $\sigma_{s,k}^2$ term captures the average local \textit{data variability}, which is controlled by the batch size $B_k$. To overcome its effect, agents must increase their batch size. Furthermore, the second term in $\sigma_s^2$ captures the \textit{model variability}, since it quantifies the suboptimality of the global model $w^o$. With Assumption \ref{assum:bdLocalMin}, we can bound its influence on the convergence of the algorithm.

\begin{theorem}[\textbf{Estimation of the second order moments of the incremental noise}]\label{thrm:incNoise}
	The incremental noise $\bm{q}_i$ defined in \eqref{eq:incNoise} has bounded variance:
	\begin{align}
		\mathbb{E}\Vert \bm{q}_i \Vert^2 &\leq O(\mu) \mathbb{E}  \Vert \widetilde{\w}_{i-1}\Vert ^2  +  O(\mu) \xi^2
	+ O( \mu^2 )\sigma_{q}^2,
	\end{align}	
	where:
		\begin{equation}
		\sigma_{q}^2 \eqdef \frac{1}{K}\sum_{k=1}^K\frac{3}{B_kN_k^2} \sum_{n=1}^{N_k} \frac{1}{p_n^{(k)}} \Vert \grad{w}Q_k(w^o; x_{k,n})\Vert^2,
	\end{equation}
	and the $O(\cdot)$ terms depend on epoch sizes, local convergence rates, total number of data samples, number of agents, Lipschitz constant, and data and agent inclusion probabilities.
%
\end{theorem}

\begin{proof}
	Proof omitted due to space limitations. 
\end{proof}
\noindent Similarly to the bound on the variance of the gradient noise, we observe a \textit{data variablity} term $\sigma_q^2$ and a \textit{model variability} term $\xi^2$. However, here the effect of the latter is greater than that of the former, since it is multiplied by an $O(\mu)$ term as opposed to $O(\mu^2)$. 

Thus, using the above results, we show under general probability sampling schemes that the ISFedAvg algorithm converges to an $O(\mu)-$neighbourhood of the global model. The result is summarized in the below theorem.

\begin{theorem}[\textbf{Mean-square-error convergence of federated learning under importance sampling}] Consider the iterates $\w_i$ generated by the importance sampling federated averaging algorithm. For sufficiently small step-size $\mu$, it holds that the mean-square-error converges exponentially fast:
	\begin{align}
		\mathbb{E}\Vert \widetilde{ \w}_i \Vert^2 \leq & O(\lambda^i) + O(\mu) \left(\sigma_s^2  + \xi^2\right) + O(\mu^3)\sigma_{q}^2, 
	\end{align}
	where $\lambda = 1- O(\mu) + O(\mu^2)  \in [0,1).$
\end{theorem}

\begin{proof}
	Proof omitted due to space limitations. 
\end{proof}
\noindent
In fact, we observe that the only significant contribution from the incremental noise is the $\xi$ term that captures the model variability accross the agents. 

\section{Importance Sampling: Probability Derivation}
Since agents and data are heterogeneous, increasing the chances of a subset being choosen is generally beneficial. In the previous section, we established a performance bound under general probabilities; the probabilities appear in the $\sigma_s^2$ term. Thus, to improve the performance bound we will choose these probabilities by minimizing $\sigma_s^2$. 
\subsection{Optimal Probabilities}
Importance sampling occurs at two levels: at the agent level when selecting the data, and at the server level when selecting the agents. During each epoch of every iteration, agent $k$ has to select a mini-batch in accordance with its normalized inclusion probabilities $p_n^{(k)}$. To calculate the optimal probabilities, we minimize $\sigma_{s,k}^2$, and we find:
\begin{equation}\label{eq:optDataProb}
	p_n^{(k),o} \eqdef \frac{\Vert\grad{w}Q_k(w^o;x_{k,n})\Vert}{\sum_{m=1}^{N_k}\Vert\grad{w}Q_k(w^o;x_{k,m})\Vert},
\end{equation}
which is proportional to the data variability. The more uniform the data is, the closer the distribution is to a uniform distribution. Similarly, during each iteration the server selects a subset of participating agents. Thus, we minimize $\sigma_s^2$, and find the optimal probabilities: 
\begin{equation}\label{eq:optAgentProb}
	p_k^o \eqdef \frac{\sqrt{\sigma_{s,k}^2 + \alpha_k \Vert \grad{w}P_k (w^o)\Vert^2}}{\sum_{\ell=1}^K \sqrt{\sigma_{s,\ell}^2 + \alpha_{\ell} \Vert \grad{w}P_{\ell} (w^o)\Vert^2}},
\end{equation}
where we define the constants $\alpha_k = \left( 3+\frac{6}{E_kB_k}\right)$. The probabilities are proportional to the two variability terms, data and model. Thus, the more homogeneous the agents are or the local data, the closer the probabilites are to a uniform distribution. Furthermore, both expressions of the probabilities are proportional to the gradient. Thus, we attribute larger probabilities to agents and samples with higher gradients resulting in a steeper gradient step during an iteration and converging faster.

\subsection{Practical Problems}
The expressions of the optimal probabilities are in terms of the true model $w^o$, which we do not know. Furthermore, they require the true gradient, which is costly to calculate. Thus, we modify the expressions by writing them in terms of the mini-batch gradient evaluated at the previous model $\w_{i-1}$. In addition, while each agent has access to all of its data and can calculate the denominator in \eqref{eq:optDataProb}, the server does not have acces to all the agents and cannot calculate the denominator in \eqref{eq:optAgentProb}. Therefore, we suggest the following scheme to calculate the probabilities: during the initial iteration we assume the normalized inclusion probability of the agents is uniform. Then, during the succeeding iterations, the server updates the probabilities of the participating agents after they have sent their current stochastic gradients:
\begin{align}\label{eq:approxPk}
	\widehat{p}_{k}^o = &\frac{\sqrt{\sigma^2_{s,k} + \alpha_k \left\Vert \widehat{\grad{w}P_k}(\bm{w}_{i-1}) \right\Vert^2}}{\sum_{\ell \in \Li} \sqrt{\sigma^2_{s,\ell} + \alpha_{\ell} \left\Vert \widehat{\grad{w}P_{\ell}}(\bm{w}_{i-1}) \right\Vert^2}}  \left( 1 - \sum_{\ell \in \Li^c} \widehat{p}_{\ell}^o \right).
	\end{align}
The multiplicative factor is to ensure all inclusion probabilities accross the agents sum to 1. Similarly, we implement the same scheme to approximate the normalized data inclusion probabilities at the agents:
\begin{align}\label{eq:approxPn}
	\widehat{p}_n^{(k),o} = &\frac{\Vert \grad{w}Q_k(\w_{i-1};x_{k,n})\Vert}{\sum_{b \in \B{k}{e}}\Vert \grad{w}Q_k(\w_{i-1};x_b)\Vert}  \left( 1 - \sum_{b \in \B{k}{e}^c}\widehat{p}_b^{(k),o}\right).
\end{align}

Given the inclusion probabilities, there are many works in the literature that devise schemes to deduce from them what the sampling probabilities should be. We choose to implement the scheme introduced in \cite{hartley1962}. If given $k=1,2,\cdots,K$ agents to choose from with normalized inclusion probabilities $p_k$, and if we require to sample $L$ agents, we first introduce the \textit{progressive totals} $\Pi_k= \sum_{\ell=1}^k L p_{\ell}$
 and set $\Pi_0 = 0$. Next, we uniformly choose at random a variable $d \in [0,1)$. Finally, we select the $L$ agents such that $ \Pi_{k-1} \leq d + i \leq \Pi_k$ for some $i \in [0,L-1] $.

\section{Experimental Section} 
\begin{figure}
	\begin{subfigure}{0.25\textwidth} 
		\centering
		\includegraphics[scale = 0.3]{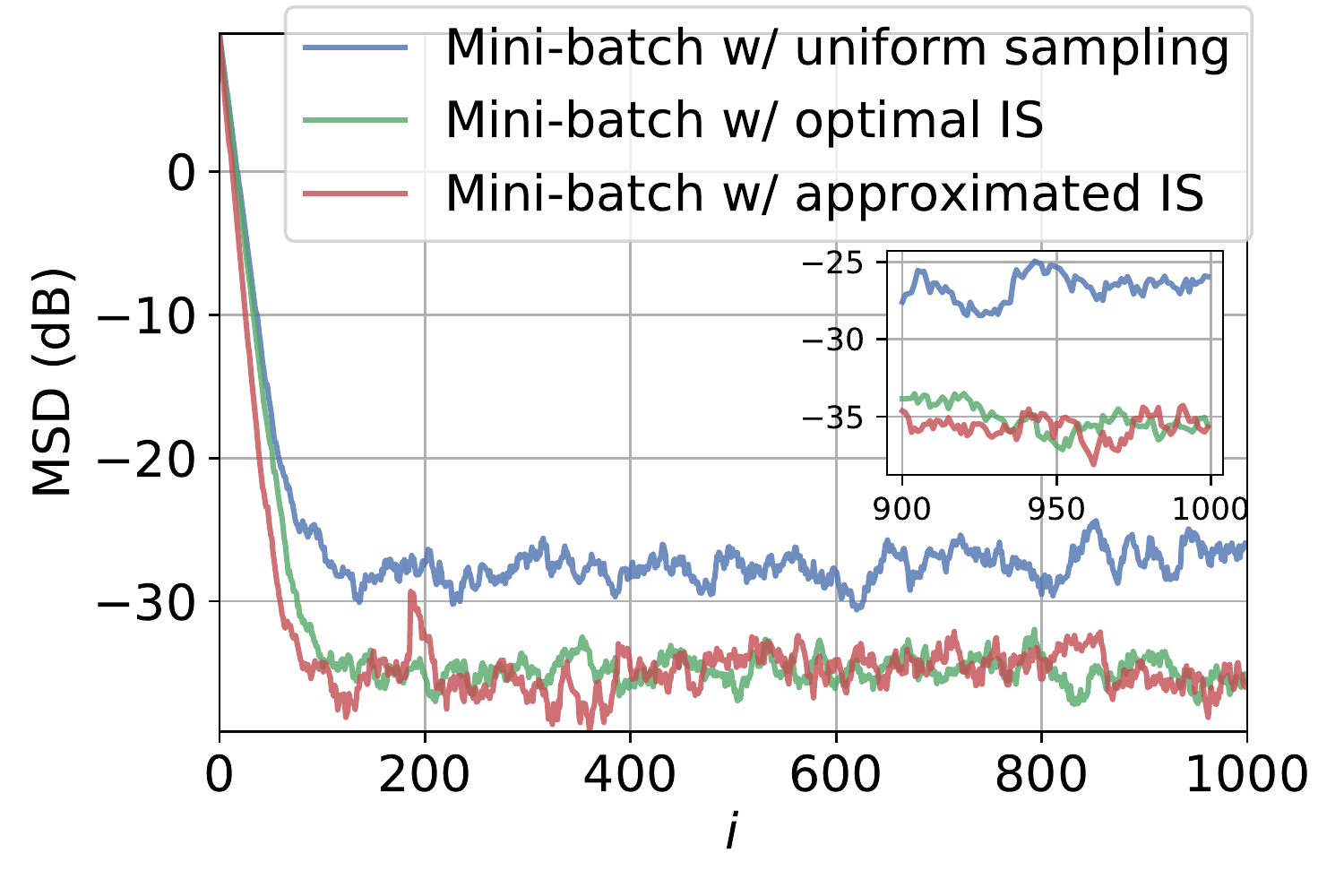}
	\end{subfigure}
\hfill
	\begin{subfigure}{0.25\textwidth} 
		\centering
		\includegraphics[scale = 0.3]{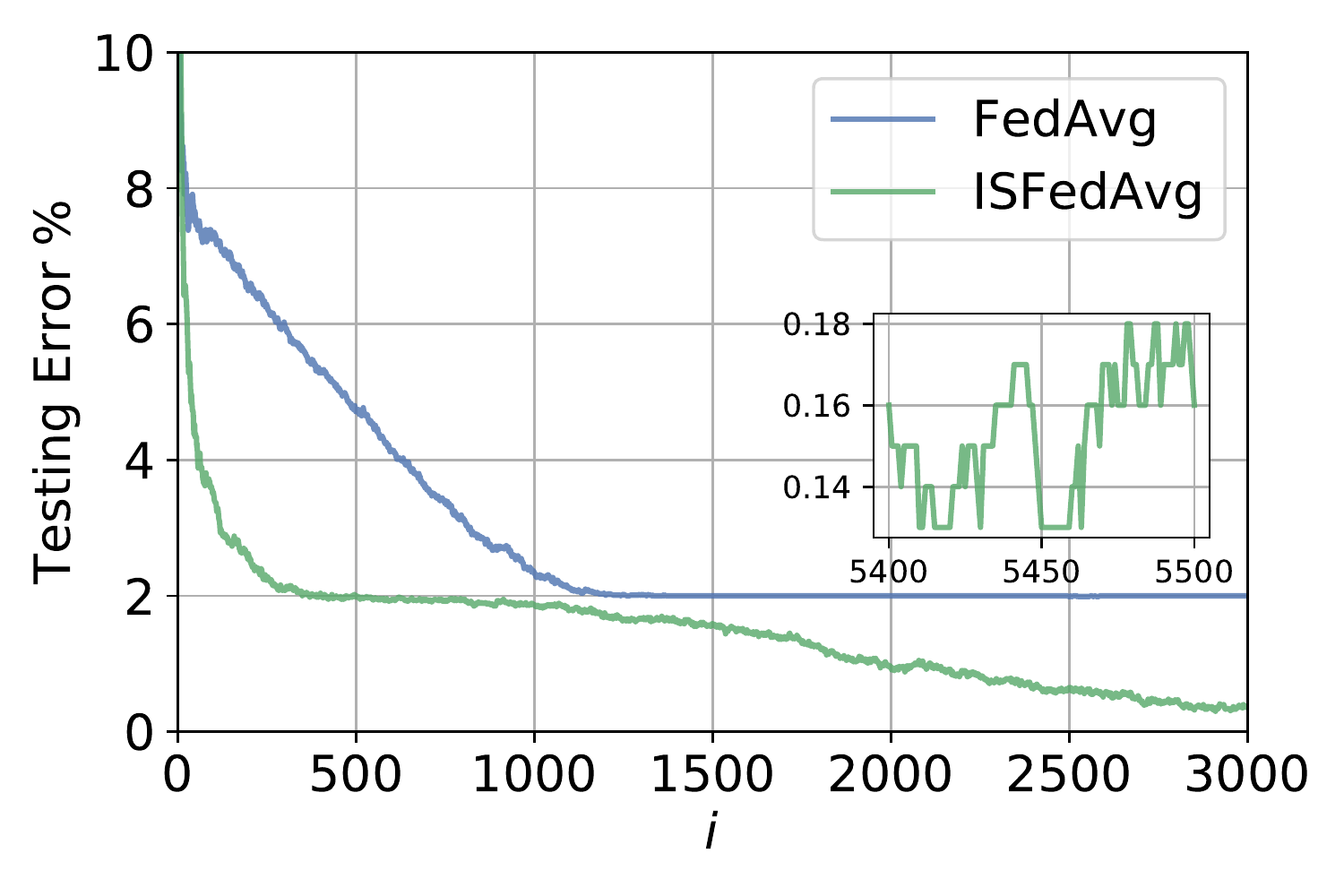}
	\end{subfigure}
\caption{Left: MSD plots of the regression problem. Right: percent testing error of the classification problem.} \label{fig:expRes}
\end{figure}

We test the algorithm on a regression problem with simulated data and a quadratic risk function, and on a classification problem with simulated data and logistic risk. Starting with the regression problem, we consider $K = 300$ agents, for which we generate $N_k = 100$ data points $\{\bm{u}_{k,n},\bm{d}_k(n)\}$ according to the following model:
\begin{equation}
	\bm{d}_k(n) = \bm{u}_{k,n} \w^{\star} + \bm{v}_k(n),
\end{equation}
where $\w^{\star}$ is some generating model and $\bm{v}_{k}(n)$ Guassian noise independent of $ \bm{u}_{k,n} $. The local risk is given by:
\begin{equation}\label{eq:LMSCost}
	P_k(w) = \frac{1}{N_k}\sum_{n=1}^{N_k} \Vert \bm{d}_k(n) - \bm{u}_{k,n}w\Vert^2 + 0.001\Vert w\Vert^2.
\end{equation} 
We set $L = 6$ and choose epoch sizes $E_k \in [1,5]$ and batch sizes $B_k \in [1,10]$ randomly. We test the performance by calculating the mean-square-deviation (MSD) at each iteration:
\begin{equation}
	\mbox{MSD} = \Vert \w_i - w^o \Vert^2,
\end{equation}
where $w^o$ is explicitly calculated for the risk \eqref{eq:LMSCost}:
\begin{align}
	w^o &= \left( \widehat{R}_{u} + \rho I\right)^{-1} \widehat{R}_u w^{\star} +\left( \widehat{R}_{u} + \rho I\right)^{-1} \widehat{r}_{uv}, \\ 
	\widehat{R}_u &\eqdef \frac{1}{K}\sum_{k=1}^K \frac{1}{N_k}\sum_{n=1}^{N_k} \bm{u}_{k,n}^\tran \bm{u}_{k,n}, \\
	\widehat{r}_{uv} &\eqdef  \frac{1}{K}\sum_{k=1}^K \frac{1}{N_k}\sum_{n=1}^{N_k} \bm{v}_k(n)\bm{u}_{k,n} . 
\end{align}
We compare 3 algorithms: FedAvg, ISFedAvg with true probabilities \eqref{eq:optAgentProb}--\eqref{eq:optDataProb}, and ISFedAvg with approximate probabilities \eqref{eq:approxPk}--\eqref{eq:approxPn}. We fix the step size $\mu = 0.01$, and run each algorithm $100$ times and average the results to get the curves in Figure \ref{fig:expRes} on the left. We observe that both importance sampling schemes outperform the standard algorithm. Furthermore, the approximate probabilities do not degrade the performance of ISFedAvg algorithm. 

We then solve a logistic regression problem on non-IID generated data $\{\bm{h}_{k,n},\gamma_k(n)\}$ for $K = 100$ agents with varying $N_k \in [20,100]$ and dimension $M = 2$. Each agent's features $\bm{h}_{k,n}$ are sampled from a normal distribution $\mathcal{N}(\mu_k,\sigma_k)$, where the variance and mean are choosen randomly. Then, the labels are calculated as follows:
\begin{equation}
	\gamma_k(n)  = \text{sign}(\bm{h}_{k,n}^\tran \bm{w}^{\star}_k),
\end{equation}
where $\bm{w}^{\star}_k$ are some random generating models that do not drift too far apart accross agents. We also generate $100$ test samples. We compare the testing error for the FedAvg and ISFedAvg algorithm, and we observe that importance sampling improves the performance from $2\%$ to almost $0\%$ (Figure \ref{fig:expRes} on the right).


\section{Conclusion}
This work introduces a two-level importance sampling scheme to federated learning: one at the level of agents and another at the level of data. The theoretical results establish convergence of the new algorithm. Optimal inclusion probabilities are derived and then approximated to overcome the practical issues. Experiments illustrate the superior performance of the proposed sampling techniques. 

%


\vfill\pagebreak


\bibliographystyle{IEEEtran}
{\balance{\bibliography{refs}}}

\end{document}